%% file: baiz.tex
\documentclass[safefonts]{baiz}
\IfFileExists{math_commands.tex}{\input{math_commands.tex}}{}
\usepackage{amsmath,amssymb,booktabs,tabularx,array,makecell,graphicx,xcolor}
\usepackage{hyperref}
\usepackage{url}
\usepackage{natbib}
\usepackage{algorithm}
\usepackage{algpseudocode}
\usepackage{capt-of}
\usepackage{subcaption}
\usepackage{wrapfig}
\usepackage{needspace}

\newcommand{\awm}{\textsc{AWM}}

\newcommand{\vjepa}{V-JEPA~2}

\newcommand{\best}[1]{\ensuremath{\mathbf{#1}}}

\newcolumntype{Y}{>{\centering\arraybackslash}X}
\newcolumntype{L}{>{\raggedright\arraybackslash}X}
\renewcommand{\arraystretch}{1.08}

\hypersetup{bookmarksdepth=3}

\begin{document}
% baiz.cls numbers \paragraph/\subparagraph as well; the manuscript uses them as
% unnumbered run-in headings (as in the ICLR version), so stop numbering there.
\setcounter{secnumdepth}{3}

% Author block: names with affiliation markers, followed by the affiliation list.
% Used as the second argument of \baiztitle (the template title box).
% 1 = University of Science and Technology of China, 2 = Baize Tongjing
% Technology Co., Ltd.  The four company-marked authors did this work while
% interning at the company (see the Acknowledgments).

% \newcommand{\baizauthors}{%
%   Ziqi Liu\textsuperscript{1,2},
%   Songhan Yang\textsuperscript{1,2},
%   Linfan Zhou\textsuperscript{1},
%   Jiatong Liu\textsuperscript{1,2},
%   Lijun Peng\textsuperscript{1,2},
%   Long Wan\textsuperscript{3},
%   Yinqi Bai%
%   \par\vspace{4pt}%
%   {\normalfont\fontsize{9}{12}\selectfont
%    \textsuperscript{1}University of Science and Technology of China (ustc.edu)\par
%    \textsuperscript{2}BAIZ Co., Ltd. (baiz.ai)\par
%    \textsuperscript{3}Harbin Institute of Technology (hit.edu.cn)\par}%
% }
\newcommand{\baizauthors}{
    BAIZ Team
}

\baiztitle
  {Abductive World Modeling via Causal Representation Learning}
  {\baizauthors}
  {\href{https://github.com/baiz-tech/AWM}{https://github.com/baiz-tech/AWM}}
  {The central challenge of world modeling is to learn representations that capture how the world evolves. However, existing world models predominantly represent future states without explicitly capturing the latent causes underlying their evolution, limiting their ability to reason about why and how the world changes. 
To address this limitation, we propose \textit{Abductive World Modeling (AWM)}, a framework that learns structured causal representations by abductively inferring latent causes from predicted futures.
Specifically, we realize AWM through the Hierarchical Abductive State Pyramid (HASP), which organizes the inferred world state into three complementary components---\textit{Entity}, \textit{Dynamic}, and \textit{Relation}---capturing what exists, how it changes, and how entities interact, respectively.
By jointly reasoning over the current observation and its predicted future, HASP abductively infers these latent factors and integrates them into a structured state representation for downstream reasoning.
% In this way, AWM shifts the focus of world modeling from merely predicting future representations to inferring the latent structure that can plausibly explain their evolution.
To the best of our knowledge, AWM is the first framework to introduce abductive state inference into latent-space world modeling for learning structured representations of world dynamics. Experiments across physical prediction, causal reasoning, and action understanding demonstrate the effectiveness of our approach. Compared with \vjepa{}, a state-of-the-art latent-space world model, AWM improves physical prediction AUROC by 10.7\%, causal reasoning accuracy by 16.8\%, and action Top-1 accuracy by 68.0\%.}
\baizmaketitlepage

\section{Introduction}

World models aim to compress observations into internal representations that enable agents to anticipate future outcomes, evaluate the consequences of different actions, and generalize across tasks\citep{ha2018worldmodels,hafner2019planet,hafner2020dreamer,
hafner2023dreamerv3,assran2025vjepa2}.
A central challenge in world modeling is therefore to learn an effective representation of the underlying world state, since the structure and quality of this representation fundamentally determine how well a model can characterize and reason about an evolving environment.

% Existing world models learn such representations through several complementary strategies. Generative and reconstruction-based approaches learn latent states by modeling the observations and dynamics required to reconstruct or generate future experience\citep{ha2018worldmodels,hafner2019planet,hafner2020dreamer,hafner2023dreamerv3,bruce2024genie}. Latent predictive methods, including JEPA-style approaches\citep{assran2023ijepa,bardes2024vjepa,zhou2025dinowm,assran2025vjepa2,wang2026internvideonext}, instead learn representations by directly predicting future states in latent space, avoiding the need to reconstruct pixel-level details. Object-centric and relational methods introduce additional structure by organizing representations around individual entities and their interactions~\citep{locatello2020slotattention,kipf2022savi,seitzer2023dinosaurs,wu2022slotformer,battaglia2016interaction,kipf2018nri,lin2020cswm}.
Existing world models learn such representations through several complementary strategies.
Generative and reconstruction-based approaches learn latent states by modeling future experience~\citep{ha2018worldmodels,hafner2019planet,hafner2020dreamer}, with later extensions scaling this paradigm to broader control and interactive environments~\citep{hafner2023dreamerv3,bruce2024genie}.
Latent predictive methods instead learn by predicting future states directly in representation space~\citep{assran2023ijepa,bardes2024vjepa,assran2025vjepa2}, including recent visual world models operating over learned features~\citep{zhou2025dinowm,wang2026internvideonext}.
Object-centric methods organize representations around individual entities~\citep{locatello2020slotattention,kipf2022savi,seitzer2023dinosaurs}, while structured dynamics models explicitly capture object interactions~\citep{wu2022slotformer,battaglia2016interaction,kipf2018nri,lin2020cswm}.

However, these representations are primarily optimized to predict what will happen, rather than to explicitly infer the latent structure that explains why and how the world evolves.
A representation can therefore accurately predict a future outcome while remaining an entangled encoding of predictive cues, without explicitly exposing the entities, dynamics, and interactions.
This limits its usefulness as a structured world state for reasoning beyond prediction itself.

To address this limitation, we propose \textbf{Abductive World Modeling (AWM)}, a world modeling framework that constructs structured representations by abductively inferring the latent factors underlying predicted world evolution.
The central principle of AWM is \emph{predict forward, then abduce backward}.
Given a current observation, a predictive video backbone first produces a latent prediction of the future.
Rather than treating this predicted future solely as a target to be matched, AWM treats it as evidence and jointly reasons over it with the current observation to infer a structured latent state.
In this way, future prediction provides evidence about what entities are present, how they are changing, and which interactions may account for the predicted evolution.
AWM therefore shifts the focus of world modeling from representing only what is likely to happen toward also recovering a structured account of the factors that explain how the world evolves.

We realize AWM through the \textbf{Hierarchical Abductive State Pyramid (HASP)}, which progressively constructs the abductive state through three specialized Attributors.
The Entity Attributor first organizes visual evidence into object-centered \textit{Entity} states, representing what exists in the scene.
Conditioned on these Entity states, the Dynamic Attributor incorporates temporal evidence to construct \textit{Dynamic} states, representing how entities change over time.
The Relation Attributor subsequently reasons over entity pairs across time to construct \textit{Relation} states, representing how entities interact.
These levels form a hierarchical abductive state that organizes predictive evidence at the entity, temporal, and interaction granularities.
By preserving lower-level information while progressively introducing higher-order structure, HASP transforms an entangled predictive representation into a structured state that can be directly examined and exploited for downstream reasoning.

% Experiments across physical prediction (Physion++), event reasoning (CLEVRER), and action understanding (EK100) show consistent gains over the \vjepa{} backbone-only baseline: the Physion++ AUROC improves by 10.7\% (from 65.20 to 72.19), CLEVRER question accuracy by 16.8\% (from 42.20 to 49.31), and EK100 action Top-1 accuracy by 68.0\% (from 25.67 to 43.12).
% Experiments across physical prediction (Physion++), event reasoning (CLEVRER), and action understanding (EK100) show consistent gains over the \vjepa{} backbone-only baseline: the Physion++ AUROC improves by 10.7\%, CLEVRER question accuracy by 16.8\%, and EK100 action Top-1 accuracy by 68.0\% .
Experiments across physical prediction (Physion++), event reasoning (CLEVRER), and action understanding (EK100) consistently outperform the \vjepa{} backbone-only baseline, improving Physion++ AUROC by 10.7\%, CLEVRER question accuracy by 16.8\%, and EK100 action Top-1 accuracy by 68.0\%. Further analyses show that Entity, Dynamic, and Relation states capture complementary object, motion, and interaction information, while targeted interventions confirm selective dependence on task-relevant Entity, temporal, and Relation evidence.

Our contributions are as follows:
\begin{itemize}
\item We introduce Abductive World Modeling (AWM), a new perspective on world-state representation that uses predicted futures as evidence to infer the latent structure underlying world evolution through a predict-forward, abduce-backward process.

\item We realize AWM through the Hierarchical Abductive State Pyramid (HASP), which hierarchically organizes predictive evidence into Entity, Dynamic, and Relation states at the object, temporal, and interaction levels. We further show that these states expose their intended factors and exhibit selective responses under targeted interventions.

% \item Experiments across physical prediction, causal reasoning, and action understanding demonstrate the effectiveness of AWM. Compared with \vjepa{}, a state-of-the-art latent-space world model, our approach improves physical prediction AUROC by 10.7\%, causal reasoning accuracy by 16.8\%, and action Top-1 accuracy by 68.0\%.
\item We extensively evaluate AWM across physical prediction, event reasoning, and action understanding, demonstrating that its structured abductive states provide consistent benefits across diverse video reasoning tasks.

\end{itemize}

\section{Related Work}

\paragraph{Latent predictive world models.}

Latent predictive world models learn representations by predicting future states rather than reconstructing future pixels. JEPA-style methods show that prediction in representation space can produce effective visual representations~\citep{assran2023ijepa,bardes2024vjepa,assran2025vjepa2}. Latent-dynamics models such as PlaNet and Dreamer support prediction, imagination, and control~\citep{hafner2019planet,hafner2020dreamer,hafner2023dreamerv3}, while recent methods model dynamics directly in visual feature spaces~\citep{zhou2025dinowm,wang2026internvideonext}. These approaches establish future prediction as an effective signal for world modeling, but their latent states are typically unified representations without explicit entity, dynamic, and relational structure.

\paragraph{Object-centric and relational representations.}

Object-centric learning organizes visual observations around individual entities.
Methods such as IODINE, Slot Attention, and DINOSAUR learn object-centered
representations~
\citep{greff2019iodine,locatello2020slotattention,seitzer2023dinosaurs},
while video extensions capture such representations over time~
\citep{kipf2022savi,zadaianchuk2023videosaur}.
Other approaches model object-centric dynamics~\citep{jiang2020scalor,wu2022slotformer,song2025ock} or explicit interactions between entities~\citep{battaglia2016interaction,kipf2018nri}. Structured dynamics models further use object-level relations to predict physical evolution~\citep{watters2017vin,lin2020cswm,sanchezgonzalez2020gns}. These methods introduce useful structure, but typically focus on individual aspects such as objects, dynamics, or interactions rather than jointly organizing all three.

\paragraph{Causal representation learning and abduction.}

Causal representation learning seeks latent variables that reflect underlying causal factors and mechanisms~\citep{scholkopf2021causal,khemakhem2020ivae}, while invariant and independent-mechanism approaches study factors that remain stable across environments~\citep{parascandolo2018independent,arjovsky2019irm}. Temporal causal representation methods further recover causal factors from sequential observations and interventions~\citep{lippe2022citris,lippe2023icitris}. Abductive reasoning instead focuses on inferring latent explanations from observations and their consequences~\citep{peirce1903abduction,zhou2019abductive}. These directions provide important foundations, but generally do not infer structured causes of world evolution from predicted states.

\section{Method}

\begin{figure}[t]
    \centering
    \includegraphics[width=0.98\linewidth]{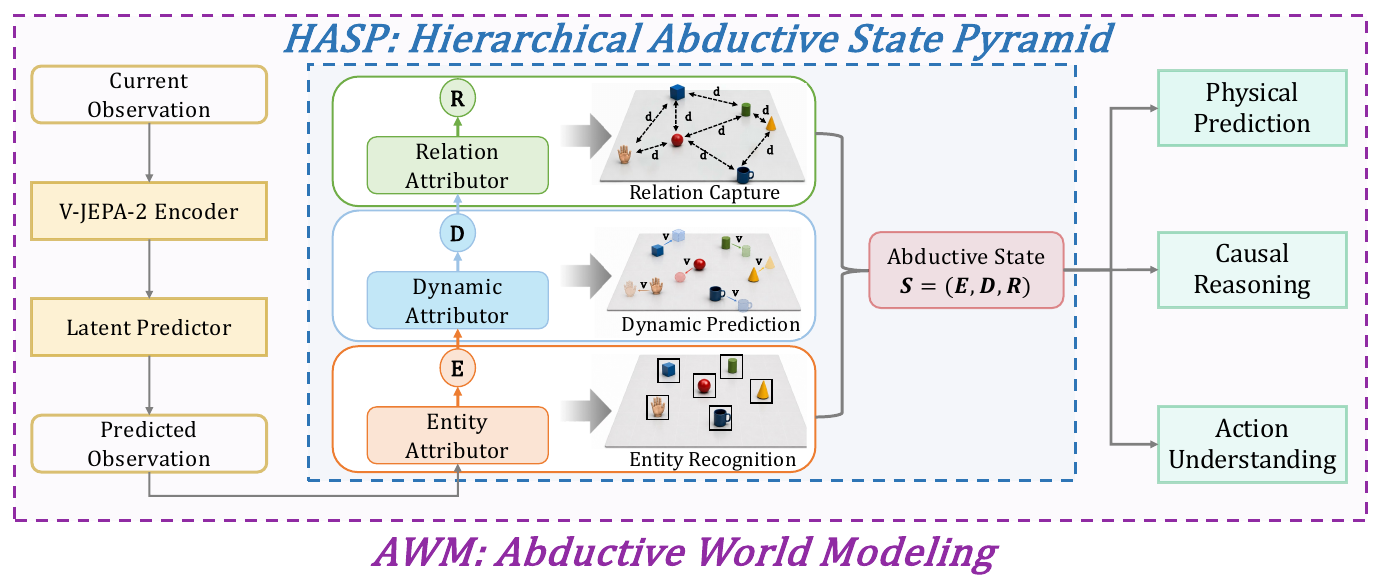}
    \caption{
    Overview of the AWM.
    Given an observed context, a frozen predictive visual backbone first predicts its future latent state.
    The HASP reasons over the current and predicted future representations to infer Entity, Dynamic, and Relation states for downstream prediction and reasoning.
    }
    \label{fig:main}
\end{figure}

We propose \textbf{Abductive World Modeling (AWM)}, which transforms predictive video representations into structured world states through a \emph{predict forward, then abduce backward} process. Given an observed context, a pretrained predictive backbone first estimates its future latent evolution, which AWM then treats as evidence for structured inference. Specifically, the \textbf{Hierarchical Abductive State Pyramid (HASP)} organizes visual evidence into \textit{Entity}, \textit{Dynamic}, and \textit{Relation} states: it extracts entities from the observed scene, attributes predicted changes to individual entities, and derives pairwise relations from their temporal evolution.

\subsection{From Future Prediction to Abductive State Inference}
Let $X_{1:t}$ denote the observed video context. A frozen predictive visual backbone consists of a visual encoder $E$ and latent predictor $P$:
\begin{equation}
    Z_c = E(X_{1:t}),
    \qquad
    \widehat{Z}_f = P(Z_c),
    \label{eq:forward-prediction}
\end{equation}
where $Z_c$ represents the observed world state and $\widehat{Z}_f$ represents its predicted future evolution in latent space. Importantly, $\widehat{Z}_f$ is generated solely from the observed context and never accesses ground-truth future frames.

Although $Z_c$ and $\widehat{Z}_f$ contain rich predictive information, they remain distributed over spatial and temporal visual tokens. Such representations indicate \emph{what future is likely}, but do not explicitly specify which entity, temporal change, or interaction accounts for that prediction. AWM therefore introduces an abductive mapping
\begin{equation}
    A:
    (Z_c,\widehat{Z}_f)
    \rightarrow
    S
    =
    \left(S^E,S^D,S^R\right),
    \label{eq:state}
\end{equation}
where $S^E$, $S^D$, and $S^R$ denote Entity, Dynamic, and Relation states.

Rather than estimating these factors independently from the same backbone representation, HASP constructs them hierarchically, progressively organizing distributed patch evidence into entities, entity-specific temporal states, and finally entity-pair temporal states. This ordering reflects the dependency structure of the reasoning problem: temporal change can only be assigned after the corresponding entity is identified, while an interaction can only be described after both entities and their temporal evolution are represented. HASP therefore performs \emph{progressive attribution}, with each level inheriting the structural units from the previous level and attributing additional predictive evidence to them.
% Rather than estimating these factors independently from the same backbone representation, HASP constructs them hierarchically:
% \begin{equation}
%     \underbrace{\text{patch}}_{\text{distributed evidence}}
%     \rightarrow
%     \underbrace{\text{entity}}_{\text{what exists}}
%     \rightarrow
%     \underbrace{\text{entity--time}}_{\text{how it changes}}
%     \rightarrow
%     \underbrace{\text{entity-pair--time}}_{\text{how entities interact}}.
%     \label{eq:granularity}
% \end{equation}
% This ordering reflects the dependency structure of the reasoning problem. Temporal change can only be assigned after the corresponding entity is identified, while an interaction can only be described after both entities and their temporal evolution are represented. HASP therefore performs \emph{progressive attribution}: each level inherits the structural units from the previous level and attributes additional predictive evidence to them.

% This ordering reflects the dependency structure of the underlying reasoning problem. Temporal change can only be assigned meaningfully after the corresponding entity has been identified, while an interaction can only be described after both participating entities and their temporal evolution have been represented. HASP therefore performs \emph{progressive attribution}: each level inherits the structural units established by the previous level and attributes additional predictive evidence to them.

\subsection{Hierarchical Abductive State Pyramid}
At a high level, HASP constructs the three states as
\begin{align}
    S^E
    &= F_E(Z_c), \\
    S^D
    &= F_D(S^E,Z_c,\widehat{Z}_f)
    + G_D(S^E,Z_c), \\
    S^R
    &= F_R(S^E,S^D)
    + G_R(S^D,Z_c),
    \label{eq:hasp}
\end{align}
where $F_E$, $F_D$, and $F_R$ are the three Attributors. $G_D$ and $G_R$ are residual information paths that preserve lower-level evidence while higher-order structure is introduced.

A critical property of this hierarchy is that the predicted future $\widehat{Z}_f$ does not directly define all three states. Entity Attribution is grounded in the observed scene, while predicted future evidence is introduced when reasoning about change in the Dynamic Attributor. Future-conditioned information is then propagated from Dynamic to Relation reasoning. Consequently, the hierarchy separates three questions: \emph{what is currently present?}, \emph{what change is implied by the predicted future?}, and \emph{which pairwise interactions are consistent with these changes?}

\subsubsection{Entity Attributor: Attributing Patch Evidence to Entities}
The first stage converts visual evidence into explicit entity-level units. Direct temporal or relational reasoning on patch tokens is undesirable because the same physical entity may occupy many spatial tokens and move across locations over time. The Entity Attributor therefore establishes a persistent object-centered coordinate system on which subsequent reasoning can operate.

We initialize $K$ learnable entity queries,
\begin{equation}
    Q_E^{(0)}
    =
    \operatorname{LearnedQueries},
\end{equation}
which attend to $Z_c$ and iteratively gather entity-specific visual evidence. At attribution layer $l$, each query updates its representation through cross-attention:
\begin{equation}
    A_E^{(l)}
    =
    \operatorname{CrossAttn}
    \left(
    \operatorname{LN}(Q_E^{(l)}),
    \operatorname{LN}(Z_c)
    \right),
\end{equation}
followed by iterative query refinement. After the final attribution layer,
\begin{equation}
    S^E = Q_E^{(L_E)}.
    \label{eq:entity-main}
\end{equation}
The resulting $S^E\in\mathbb{R}^{K\times d}$ replaces the original patch organization with a fixed set of object-centered states. Conceptually, this stage answers the first abductive question:
\begin{center}
    \emph{Which parts of the distributed visual evidence can be attributed to the same entity?}
\end{center}
Only the current representation $Z_c$ is required at this stage because identifying what exists should not depend on hallucinating an entity from a predicted future. This also establishes a stable entity basis before future-dependent reasoning is introduced.

\subsubsection{Dynamic Attributor: Attributing Predicted Change to Entities}
Entity states alone characterize \emph{what exists}, but do not explain how individual entities account for the predicted transition. The Dynamic Attributor therefore lifts every entity into an entity-specific temporal representation and attributes current--future differences to that entity.

For each Entity state, we construct temporal queries
\begin{equation}
    Q_D^{(0)}
    =
    \phi_E(S^E)
    + E_{\mathrm{time}}
    + E_{\mathrm{future}},
    \label{eq:dynamic-query-main}
\end{equation}
where $\phi_E(S^E)$ preserves entity identity, $E_{\mathrm{time}}$ specifies temporal position, and $E_{\mathrm{future}}$ explicitly distinguishes predicted-future positions from observed ones.

These entity-conditioned queries jointly attend to the current and predicted future evidence:
\begin{equation}
    H_D
    =
    \operatorname{CrossAttn}
    \left(
    Q_D^{(0)},
    [Z_c;\widehat{Z}_f]
    \right).
    \label{eq:dynamic-attn-main}
\end{equation}
Temporal dependencies are subsequently integrated along each entity trajectory,
\begin{equation}
    S^D
    =
    \operatorname{TemporalTransformer}(H_D)
    + G_D(S^E,Z_c).
    \label{eq:dynamic-main}
\end{equation}

% This design is central to the abductive interpretation of AWM. The predicted future does not simply provide additional visual features. Instead, it acts as evidence about \emph{what must change} between the observed and expected future states. Because the temporal queries are initialized from $S^E$, this evidence is attributed separately to individual entities rather than being absorbed into a global video representation.
This design is central to AWM's abductive interpretation. The predicted future provides evidence of what must change, while temporal queries initialized from $S^E$ preserve entity-specific attribution.

The main attribution path therefore captures future-conditioned temporal change, whereas $G_D$ retains entity and current-state information that should not be discarded when modeling motion. The resulting $S^D$ is organized at the entity--time level and answers the second abductive question:
\begin{center}
    \emph{Given an entity, what temporal change is supported by the observed and predicted future states?}
\end{center}
This is also the point at which predicted-future evidence first enters the hierarchy. As a result, HASP uses the future specifically to explain dynamics rather than to redefine entity identity.

\subsubsection{Relation Attributor: Attributing Dynamics to Interactions}
Changes of individual entities are still insufficient to represent many world dynamics. Collision, contact, manipulation, pursuit, and other interactions are inherently defined between entities. The Relation Attributor therefore converts entity-specific dynamics into explicit pairwise temporal states.

For each candidate entity pair $(i,j)$ and temporal position $t$, we construct
\begin{equation}
    r_{ij,t}^{(0)}
    =
    \left[
    S_i^E + S_j^E,\;
    |S_i^E-S_j^E|,\;
    S_{i,t}^D + S_{j,t}^D,\;
    |S_{i,t}^D - S_{j,t}^D|
    \right].
    \label{eq:relation-input-main}
\end{equation}
The shared terms describe properties jointly expressed by the pair, whereas the difference terms explicitly capture their relative entity and dynamic states. This representation is subsequently projected and temporally aggregated:
\begin{equation}
    S^R_{ij,1:T}
    =
    \operatorname{TemporalAttention}
    \left(
    \operatorname{MLP}_R(r_{ij,1:T}^{(0)})
    \right)
    + G_R(S^D,Z_c).
    \label{eq:relation-main}
\end{equation}
Relation reasoning is deliberately performed after Dynamic attribution. Instead of attempting to infer interactions directly from raw visual tokens, each relation is constructed from two already-grounded entities and their entity-specific temporal evolution. Thus, interaction evidence is represented in a pair-specific coordinate system.

Because $S^D$ already incorporates evidence from $\widehat{Z}_f$, future information is propagated naturally into Relation inference without requiring the Relation Attributor to independently decode the predicted future representation. The resulting $S^R$ is organized at the entity-pair--time level and answers the third abductive question:
\begin{center}
    \emph{Which interaction between two entities can account for their attributed temporal evolution?}
\end{center}
The explicit pair organization additionally makes individual relations addressable. A particular $(i,j)$ state can therefore be independently probed, masked, or intervened upon, which is important for the factor-specific analyses introduced in Sec.~\ref{sec:analysis}.

\subsection{Training Objective}
HASP is trained with supervision aligned to the native granularity of each state:
\begin{equation}
    \mathcal{L}_{\mathrm{HASP}}
    =
    \mathcal{L}_E
    + \mathcal{L}_D
    + \mathcal{L}_R
    + \lambda_{\mathrm{task}}\mathcal{L}_{\mathrm{task}},
    \label{eq:hasp-loss}
\end{equation}
where $\mathcal{L}_E$, $\mathcal{L}_D$, and $\mathcal{L}_R$ supervise Entity, Dynamic, and Relation states at the object, entity-time, and pairwise interaction levels, respectively, while $\mathcal{L}_{\mathrm{task}}$ provides optional downstream task supervision.

This factor-aligned supervision is important because HASP is not intended merely to increase representation capacity. Instead, each level is encouraged to expose information at the structural granularity it is designed to represent. Entity supervision encourages object-centered attribution, Dynamic supervision encourages entity-specific temporal attribution, and Relation supervision encourages pair-specific interaction attribution.

Throughout training, the pretrained encoder $E$ and latent predictor $P$ remain frozen, while only HASP and the task-specific readouts are optimized. This setup isolates the contribution of abductive state construction: any performance gains come from reorganizing the predictive evidence already provided by the backbone into structured Entity, Dynamic, and Relation states, rather than from further adapting or improving the underlying future predictor.
% Additional architectural dimensions, iterative attribution details, and benchmark-specific optimization settings are provided in Appendix~\ref{app:hasp-detail} and Appendix~\ref{app:training-detail}.
Additional tensor dimensions and HASP implementation details are provided in Appendix~\ref{app:backbone-detail} and Appendix~\ref{app:hasp-detail}, while factor-specific training objectives are described in Appendix~\ref{app:training-detail}.

\section{Results}

We evaluate AWM from three complementary aspects.
\textbf{Experiment 1} compares its downstream performance with baselines across physical prediction, event reasoning, and action recognition.
\textbf{Experiment 2} analyzes whether the Entity, Dynamic, and Relation states capture the object, motion, and interaction information associated with their respective Attributors.
\textbf{Experiment 3} tests whether predictions depend selectively on the corresponding object, temporal, and relational evidence.

\textbf{Tasks and Datasets.}
We evaluate AWM on three complementary video benchmarks: Physion++~\citep{bear2021physion} for physical prediction, CLEVRER~\citep{yi2020clevrer} for future-event and causal reasoning, and EPIC-KITCHENS-100 (EK100)~\citep{damend2022epic} for fine-grained action recognition. Dataset details are provided in Appendix~\ref{app:dataset_detail}.

\textbf{Experimental Setup.}
AWM is built on the V-JEPA 2 predictive video backbone, with HASP producing Entity, Dynamic, and Relation states. The backbone and HASP are frozen during evaluation, and only lightweight probes or task-specific readouts are trained. 
% All matched comparisons use the same data splits, preprocessing, readout capacity, optimization, and training budget. Experiments are conducted on eight NVIDIA A800 80GB GPUs; further details are provided in Appendix~\ref{appendix:exp_setup}.
% All matched comparisons use the same data splits and evaluation settings.
Experiments are conducted on 32 NVIDIA A800 80GB GPUs. Additional implementation and baseline evaluation details are provided in Appendix~\ref{appendix:exp_setup}.

\textbf{Baselines.}
We compare AWM with V-JEPA 2~\citep{assran2025vjepa2}, Orca~\citep{wang2026orca}, and VideoMAE~v2~\citep{wang2023videomaev2}. V-JEPA 2 serves as the matched predictive baseline, sharing the same backbone and input protocol with AWM, while Orca and VideoMAE~v2 provide external video representation baselines. All methods are evaluated under the same task-specific protocol whenever applicable.More details are provided in Appendix~\ref{appendix:exp_setup}.

\input{Table/main_results}

\subsection{Experiment 1. Main Results}

In this subsection, we compare AWM with V-JEPA 2, Orca, and VideoMAE~v2 across physical prediction, event reasoning, and egocentric action recognition. As shown in Table~\ref{tab:main}, AWM achieves the best overall performance across all evaluated task spaces. Relative to the matched V-JEPA 2 baseline, AWM improves Physion++ AUROC by 10.7\%, balanced accuracy by 7.5\%, and accuracy by 6.5\%, outperforming all three baselines. On CLEVRER, AWM improves option accuracy by 8.3\% and question accuracy by 16.8\% over V-JEPA 2. Orca and VideoMAE~v2 are not evaluated on CLEVRER because their released models do not provide a predictive module required by the CLEVRER evaluation pipeline.
On EK100, AWM improves verb Top-1/Top-5 accuracy by 23.9\%/5.1\%, noun Top-1/Top-5 accuracy by 37.3\%/19.8\%, and action Top-1/Top-5 accuracy by 68.0\%/43.5\% over V-JEPA 2, consistently outperforming all baselines. These results demonstrate that the proposed state interface provides strong and consistent gains across heterogeneous video understanding tasks. The following experiments further analyze whether these improvements can be attributed to the intended Entity, Dynamic, and Relation structure.

\begin{figure*}[t]
\centering

\begin{subfigure}[t]{0.31\textwidth}
\centering
\includegraphics[width=\linewidth]{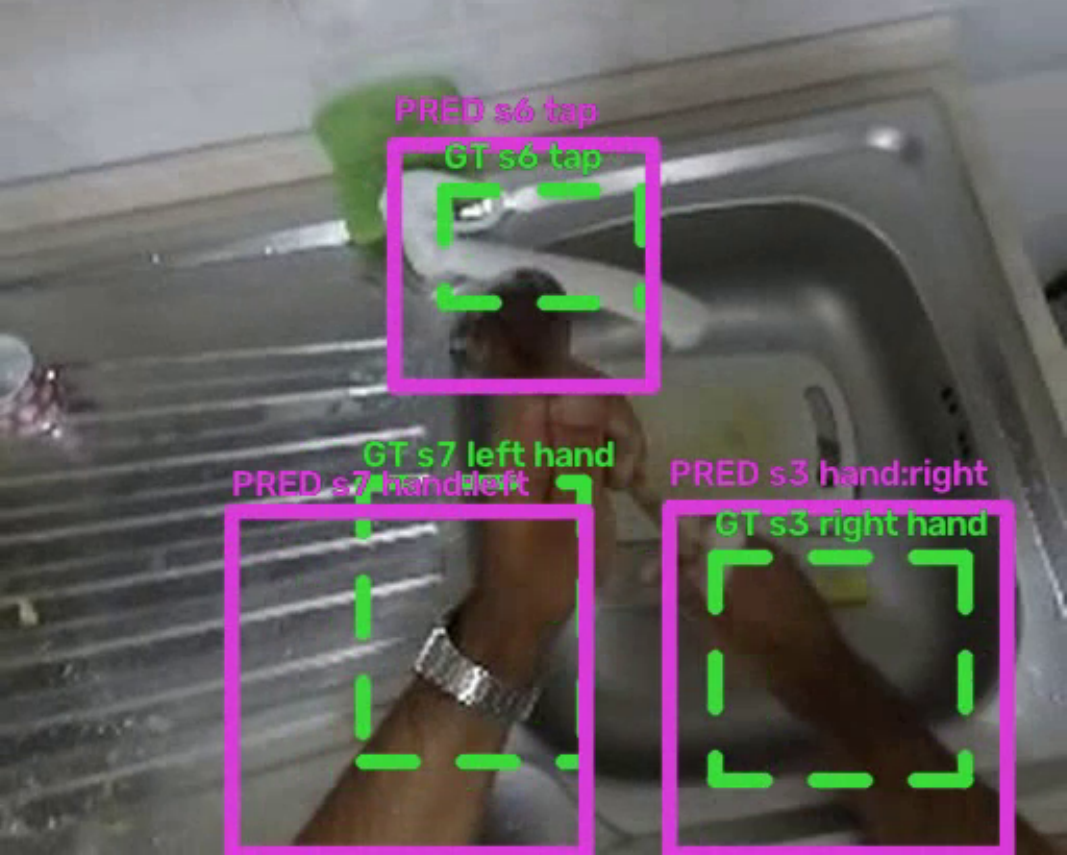}
\caption{Egocentric Entity grounding.}
\label{fig:entity-grounding-ego1}
\end{subfigure}
\hfill
\begin{subfigure}[t]{0.31\textwidth}
\centering
\includegraphics[width=\linewidth]{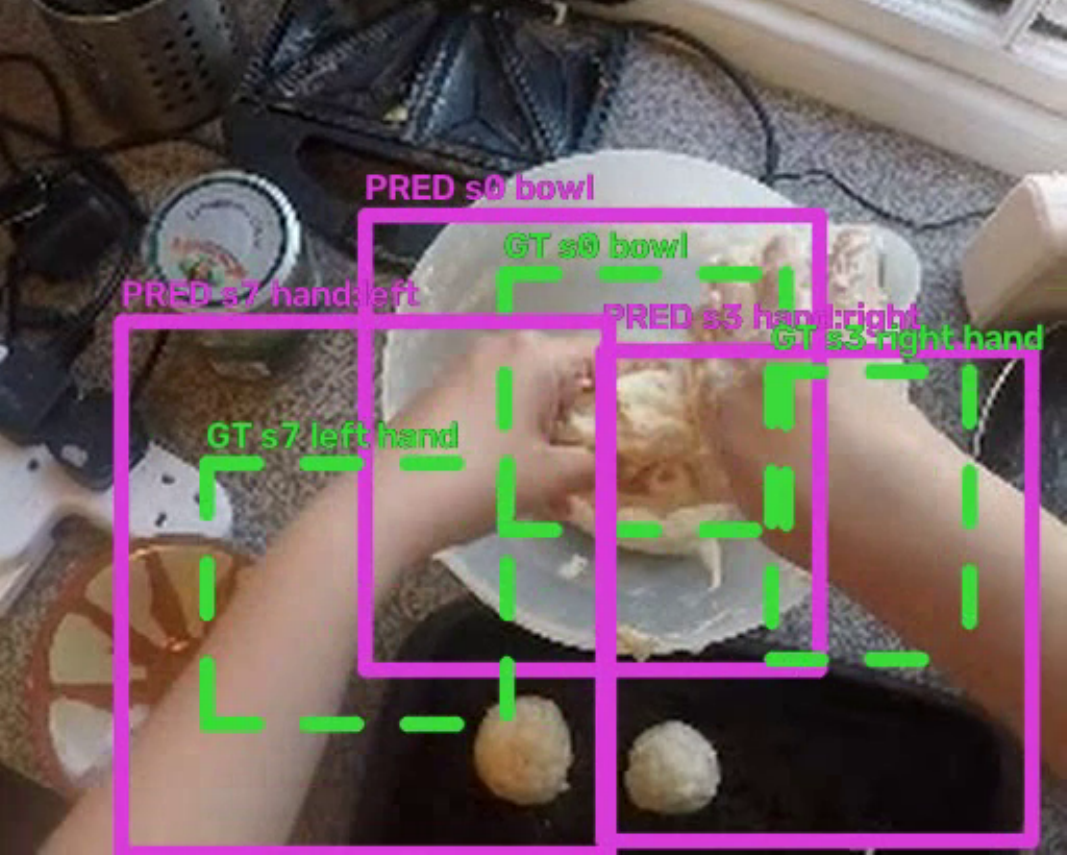}
\caption{Egocentric Entity grounding.}
\label{fig:entity-grounding-ego2}
\end{subfigure}
\hfill
\begin{subfigure}[t]{0.31\textwidth}
\centering
\includegraphics[width=\linewidth]{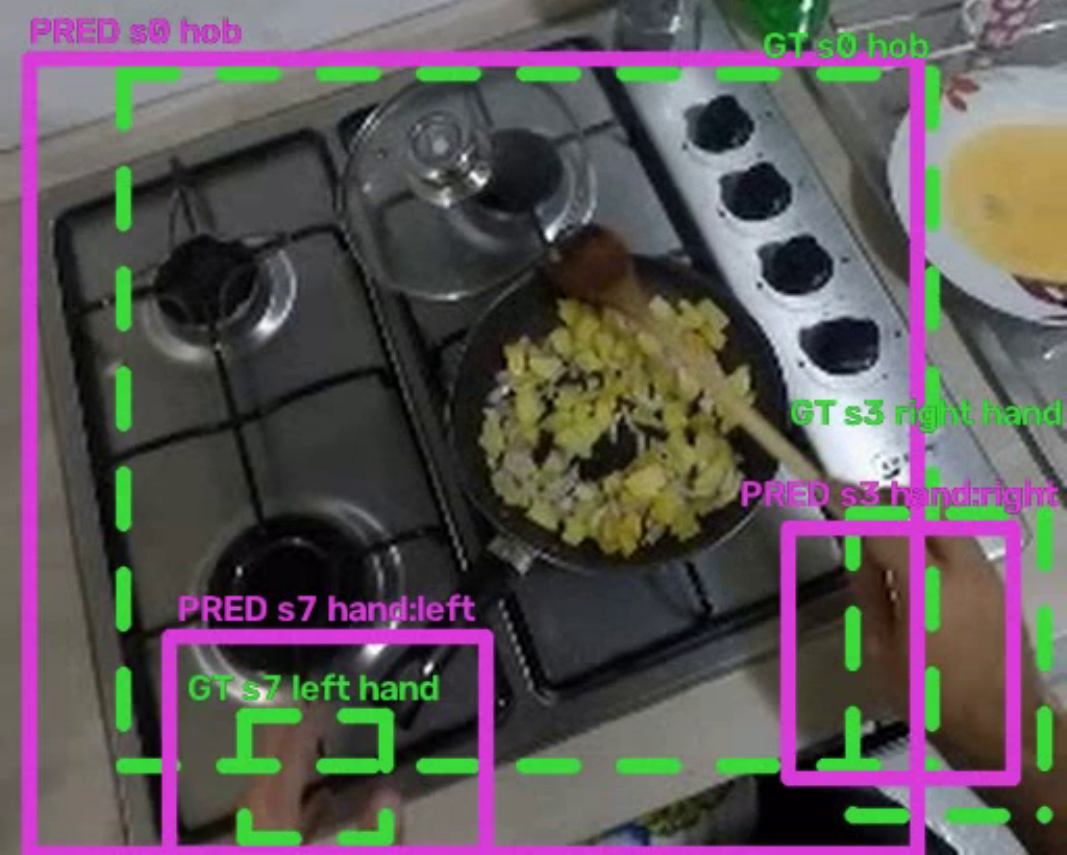}
\caption{Egocentric Entity grounding.}
\label{fig:entity-grounding-ego3}
\end{subfigure}

\vspace{4pt}

\begin{subfigure}[t]{0.98\textwidth}
\centering
\includegraphics[width=\linewidth]{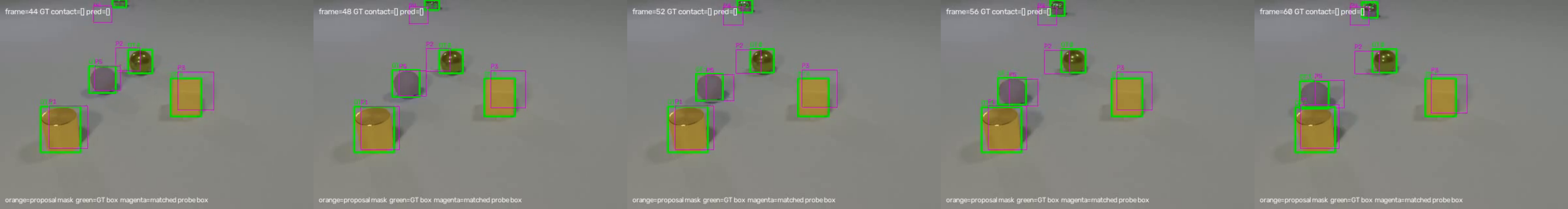}
\caption{Temporal grounding across consecutive frames.}
\label{fig:entity-grounding-temporal}
\end{subfigure}

\caption{
Qualitative visualization of Entity grounding on EK100 and CLEVRER.
(a)--(c) show representative EK100 scenes, where predicted Entity regions align with ground-truth hands and manipulated objects.
(d) shows consecutive CLEVRER frames, where matched Entity predictions remain aligned with the corresponding ground-truth objects over time.
Green boxes denote ground-truth regions and magenta boxes denote matched predicted regions.
}
\label{fig:entity-grounding}

\end{figure*}

\subsection{Experiment 2. Analysis of Attributor Interpretability}
\label{sec:analysis}

We analyze whether the Entity, Dynamic, and Relation states expose the object, temporal, and interaction information associated with their respective Attributors.

\paragraph{Native Factor Readability.}

\begin{wraptable}{r}{0.42\linewidth}
\vspace{-8pt}
\centering
\caption{
Native factor readability on Physion++.
}
\label{tab:native-factor-information}

\scriptsize
\renewcommand{\arraystretch}{1.06}
\setlength{\tabcolsep}{3.5pt}

\begin{tabular*}{\linewidth}{
    @{\extracolsep{\fill}}
    l
    l
    c
    c
}
\toprule\toprule
\textbf{State}
& \textbf{Factor}
& \textbf{Metric}
& \textbf{Score} \\
\midrule

Entity
& Presence
& AUROC
& 0.9634 \\

& Spatial extent
& $R^2$
& 0.5316 \\

\addlinespace[1.5pt]

Dynamic
& Speed
& $R^2$
& 0.7198 \\

& Signed velocity
& $R^2$
& 0.3805 \\

\addlinespace[1.5pt]

Relation
& Contact
& AUROC
& 0.9789 \\

& Pairwise distance
& $R^2$
& 0.5815 \\

& Time-to-contact
& $R^2$
& 0.5542 \\

\bottomrule
\end{tabular*}

\vspace{-6pt}
\end{wraptable}

% We evaluate the three state levels on Physion++ using probes defined at their native representation granularities. Entity slots are probed for object presence and spatial extent, Dynamic slot-time states for speed and signed velocity, and Relation pair-time states for distance, contact, and time-to-contact (TTC).
% As shown in Table~\ref{tab:native-factor-information}, Entity provides strong object-level information, achieving 0.9634 AUROC for presence and $R^2=0.5316$ for spatial extent. Dynamic captures temporal information, with $R^2=0.7198$ for speed and $R^2=0.3805$ for signed velocity.
% Relation provides strong interaction information, achieving 0.9789 AUROC for contact while also supporting distance and TTC prediction.
% These results show that the three state levels expose complementary information at their intended entity, temporal, and pairwise granularities. Additional slot-level object identity diagnostics are provided in Appendix~\ref{app:attributor-analysis}.

We evaluate the three state levels on Physion++ at their native granularities: Entity for object-level factors, Dynamic for motion, and Relation for pairwise interactions.
As shown in Table~\ref{tab:native-factor-information}, Entity strongly captures object presence and spatial extent, Dynamic captures motion factors, and Relation achieves high contact readability while retaining distance and time-to-contact information.
Together, these results show that the three state levels expose complementary information at their intended entity, temporal, and pairwise granularities.
Additional analyses are provided in Appendix~\ref{app:attributor-analysis}.

\paragraph{Qualitative Entity Grounding.}

Beyond factor-level probes, we qualitatively examine whether Entity states capture localized and temporally coherent object information.
As shown in Figure~\ref{fig:entity-grounding}, predicted Entity regions align with ground-truth objects in EK100, including hands and manipulated objects, while remaining consistently aligned with the same objects across consecutive CLEVRER frames.
These results provide qualitative evidence that Entity states preserve object-level grounding across both real and synthetic scenes.

% \Needspace{0.3\textheight}

\begin{wrapfigure}{r}{0.44\columnwidth}
    \centering
    \includegraphics[
        width=\linewidth,
        keepaspectratio
    ]{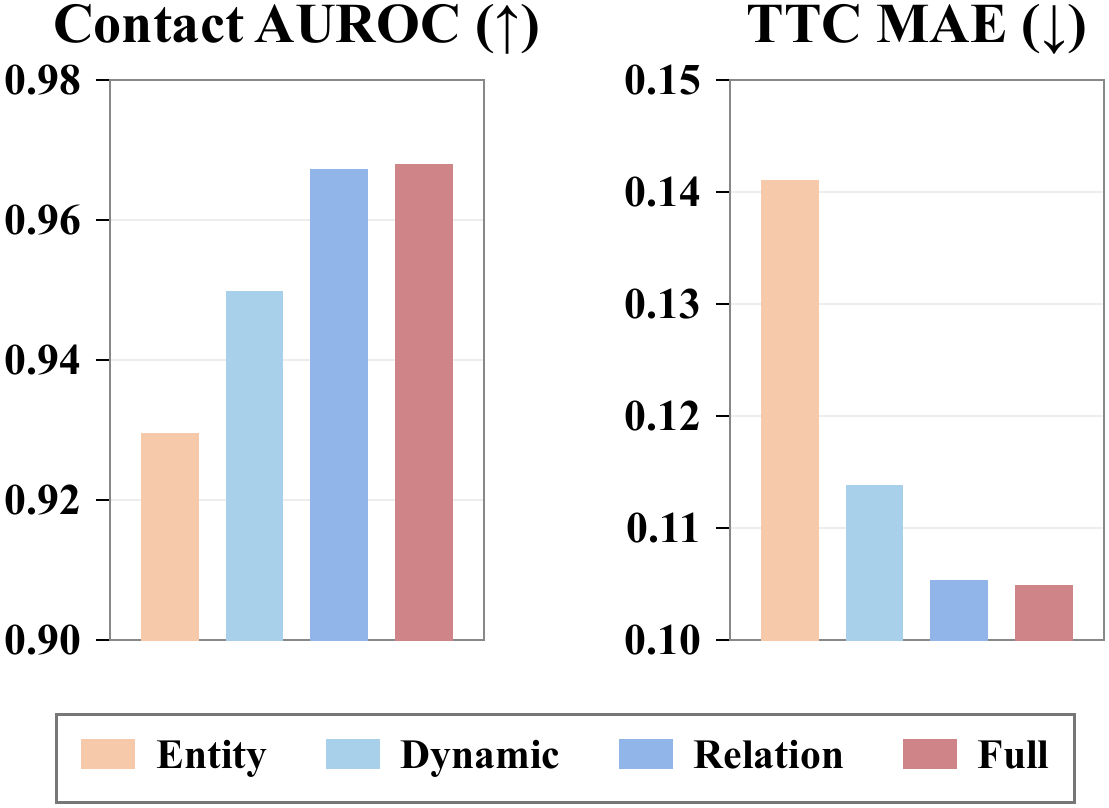}
    \caption{
        Interaction prediction with individual HASP states on CLEVRER.
        \textit{Full} uses all three states.
    }
    \label{fig:relation-component}
\end{wrapfigure}

\paragraph{Relation Interaction.}
We examine whether interaction information is specifically concentrated in the Relation state.
On CLEVRER, we compare task readouts based on Entity, Dynamic, Relation, and the full state representation for contact prediction and TTC estimation.
As shown in Figure~\ref{fig:relation-component}, Relation performs close to the full representation on both interaction-related tasks, while Entity and Dynamic are less effective when used alone. This suggests that most pairwise interaction information is already captured at the Relation level, whereas the lower-level states primarily encode complementary object and motion information. The result is consistent with the hierarchical design of HASP, where Entity establishes object-level structure, Dynamic introduces entity-specific temporal evolution, and Relation integrates these cues into explicit pairwise interaction states.

\WFclear

\subsection{Experiment 3. Ablation Study on Intervention Consistency}

The previous analysis shows that Entity, Dynamic, and Relation states expose readable object, motion, and interaction information. We further test whether downstream predictions depend selectively on the corresponding evidence. 

\paragraph{Entity Intervention.}

\begin{figure*}[t]
    \centering
    \includegraphics[
        width=0.8\textwidth,
        keepaspectratio
    ]{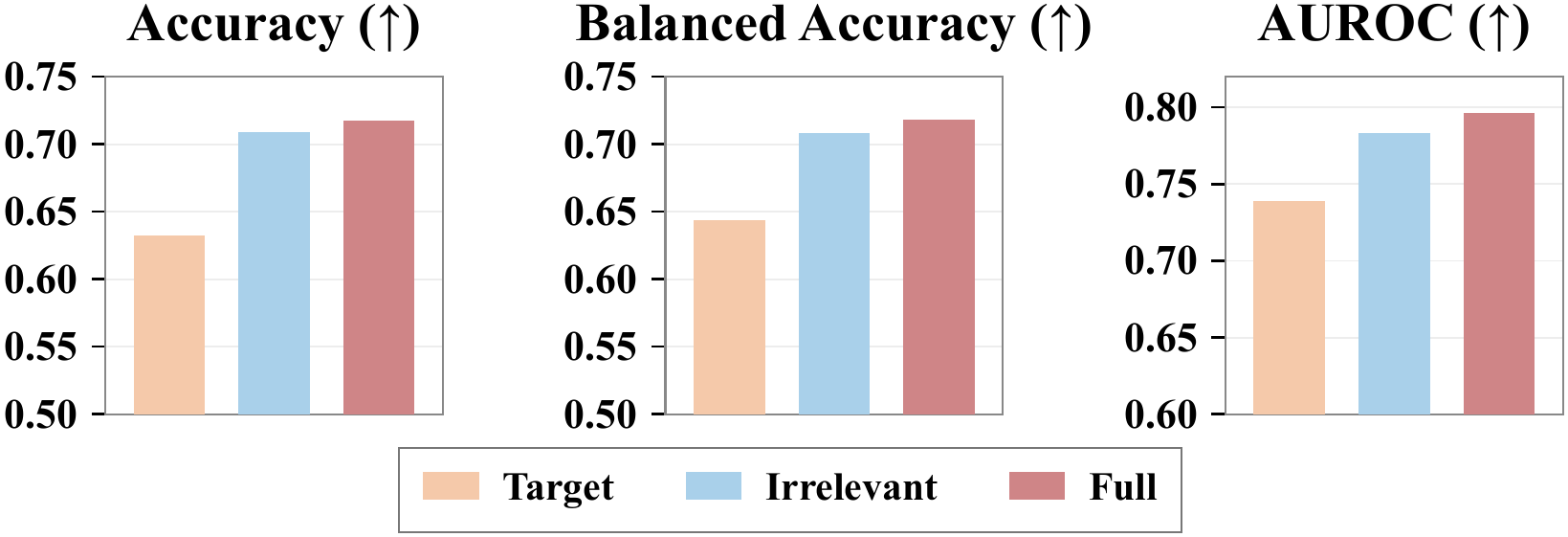}

    \caption{
    Entity intervention on Physion++ OCP prediction.
    Target masks the target Entity slot,
    Irrelevant masks a non-target slot,
    and Full retains all Entity slots.
    Performance is evaluated by accuracy, balanced accuracy, and AUROC ($\uparrow$),
    showing how prediction quality changes when target-relevant or irrelevant entity information is removed.
    }
    \label{fig:entity-intervention}
\end{figure*}

We test whether physical outcome prediction on Physion++ depends selectively on object-level evidence represented by the Entity state.
As shown in Figure~\ref{fig:entity-intervention}, masking the target-object slot reduces OCP accuracy from 0.7175 to 0.6325 and AUROC from 0.7962 to 0.7391. In contrast, masking an irrelevant-object slot retains substantially higher performance, with 0.7088 accuracy and 0.7834 AUROC. The larger degradation under target-object masking indicates that downstream prediction depends more strongly on task-relevant Entity evidence than on arbitrary object information.
Additional confusion-matrix and paired-logit analyses of this intervention are reported in Appendix~\ref{app:entity-intervention-diagnostics}.

\paragraph{Dynamic Intervention.}

\input{Table/input_temporal_intervention}

We examine whether temporal variation in the predictive representation is functionally important for Dynamic reasoning. We construct a temporal-static input by averaging the context and predicted future features over time and repeating the resulting features at every temporal position, while keeping all model components and readouts fixed.
As shown in Table~\ref{tab:input-temporal-intervention}, removing temporal variation increases Speed MAE from 0.0670 to 0.1120, decreases Contact AUROC from 0.9837 to 0.9485, and reduces OCP AUROC from 0.7907 to 0.5646. The consistent degradation across motion, interaction, and physical outcome prediction shows that temporal variation provides important evidence for downstream reasoning.

\paragraph{Relation Intervention.}

\input{Table/relation_selective_intervention}

We test whether interaction prediction depends selectively on the Relation state associated with the relevant entity pair. On CLEVRER, we compare masking the interacting pair with masking a non-contact pair.
As shown in Table~\ref{tab:relation-selective-intervention}, masking the relevant contact pair reduces Contact AUROC from 0.9633 to 0.9488 and increases TTC MAE from 0.1306 to 0.1376. Masking a non-contact pair causes a smaller change, yielding 0.9518 Contact AUROC and 0.1349 TTC MAE. This stronger sensitivity to the interacting pair indicates that Relation states capture pair-specific information that is directly used for contact and TTC prediction.

Together, these interventions provide functional evidence for the three levels of HASP: downstream predictions selectively depend on task-relevant Entity evidence, temporal variation associated with Dynamic reasoning, and interaction-specific Relation states. Additional query-conditioned object intervention results on CLEVRER are provided in Appendix~\ref{app:query-object-intervention}.

\section{Conclusion}

Existing world models are effective at predicting future states, but their representations often lack explicit causal structure for explaining how the world evolves. In this paper, we proposed \textbf{Abductive World Modeling (AWM)}, which follows the principle of \emph{predict forward, then abduce backward} to infer latent causes from predicted future states. AWM instantiates this process with the Hierarchical Abductive State Pyramid (HASP), which organizes world dynamics into complementary Entity, Dynamic, and Relation factors that capture what exists, how it changes, and how entities interact. Experiments across physical prediction, causal reasoning, and action understanding demonstrate the effectiveness of the resulting structured causal representations, improving over the V-JEPA 2 backbone by 10.7\% in Physion++ AUROC, 16.8\% in CLEVRER question accuracy, and 68.0\% in EK100 action Top-1 accuracy.

\subsection*{AI Use Statement}
Generative AI tools were used to assist with literature organization and editorial drafting.  All technical claims, equations, experimental numbers, and code references were checked against the project files by the authors, who take responsibility for the final manuscript.

\subsection*{Reproducibility Statement}
The source code, dataset-isolated training protocols and analysis scripts will be
released upon publication. The analysis scripts export reusable state features and
reproduce the factor probes and intervention tables without modifying training
checkpoints; model checkpoints are not released.
\pdfbookmark[1]{Acknowledgments}{ack}
\subsection*{Acknowledgments}
% Chinese registered name: 白泽通境科技有限公司
This work was conducted at and supported by Baize Tongjing Technology Co., Ltd.
We thank <names> for data preparation, engineering support, and helpful
discussions.
Experiments were run on 32$\times$ NVIDIA A800 80GB GPUs provided by the company.

\pdfbookmark[1]{Funding, Compute, and Compliance}{compliance}
\subsection*{Funding, Compute, and Compliance}
\textbf{Funding.} This work was funded by Baize Tongjing Technology Co., Ltd.
under project <internal project id>.

\textbf{Compute.} All experiments were conducted on 32$\times$ NVIDIA A800 80GB
GPUs provided by the company.

\textbf{Author status.} Ziqi Liu, Songhan Yang, Jiatong Liu and Lijun Peng
contributed to this work while interning at Baize Tongjing Technology Co., Ltd.
Their \texttt{.edu.cn} addresses are personal contact addresses and do not
represent their home institutions.

\textbf{Publication review.} This manuscript has been reviewed and approved for
external publication under the company's research publication policy, and
contains no confidential, customer-identifying or export-controlled information.

\textbf{Intellectual property.} The methods described here were developed as
part of the authors' work at the company, which retains rights to the associated
model, code and checkpoints.

\textbf{Datasets.} We use Physion++~\citep{bear2021physion},
CLEVRER~\citep{yi2020clevrer} and EPIC-KITCHENS-100~\citep{damend2022epic}
strictly under their original licenses and terms of use, for research purposes
only. No new human-subject data was collected.

\textbf{Corresponding author.} <corresponding author>, \texttt{<email>}. The views
expressed in this paper are those of the authors and do not necessarily reflect
those of the company.

\bibliography{baiz}
\bibliographystyle{baiz}

\appendix

\section{Details of Method}
\label{app:method-detail}

This appendix provides implementation-level details of the predictive backbone interface and the three Attributors in HASP. The description follows the formulation in the main paper and focuses on the tensor organization and computational flow.

\subsection{Predictive Backbone Interface}
\label{app:backbone-detail}

Given an observed video context $X_{1:t}$, the frozen visual encoder produces
\begin{equation}
    Z_c = E(X_{1:t}),
    \qquad
    Z_c
    \in
    \mathbb{R}^{B \times T_c \times N \times d_v},
    \label{eq:app-encoder}
\end{equation}
where $B$ denotes the batch size, $T_c$ the number of observed temporal positions, $N$ the number of spatial visual tokens, and $d_v$ the backbone feature dimension.

The frozen latent predictor subsequently produces
\begin{equation}
    \widehat Z_f = P(Z_c),
    \qquad
    \widehat Z_f
    \in
    \mathbb{R}^{B \times T_f \times N \times d_v},
    \label{eq:app-predictor}
\end{equation}
where $T_f$ denotes the number of predicted future positions. No ground-truth future representation is used to construct $\widehat Z_f$.

For $K$ entity slots and state dimension $d$, HASP produces
\begin{equation}
    S^E
    \in
    \mathbb{R}^{B\times K\times d},
\end{equation}
\begin{equation}
    S^D
    \in
    \mathbb{R}^{B\times K\times T_D\times d},
\end{equation}
and
\begin{equation}
    S^R
    \in
    \mathbb{R}^{B\times |\mathcal P|\times T_R\times d},
\end{equation}
where
\begin{equation}
    \mathcal P
    =
    \{(i,j)\mid 1\leq i<j\leq K\}
\end{equation}
is the set of unordered candidate entity pairs.

In the Physion++ implementation, $K=8$ and $d=256$, yielding
\begin{equation}
    |\mathcal P|=\binom{8}{2}=28.
\end{equation}

\subsection{Implementation of HASP}
\label{app:hasp-detail}

\subsubsection{Entity Attributor}

The Entity Attributor converts the current patch-organized representation into a fixed set of entity-organized states. We initialize $K$ learnable entity queries
\begin{equation}
    Q_E^{(0)}
    \in
    \mathbb{R}^{K\times d}.
    \label{eq:app-entity-query}
\end{equation}

The current representation is flattened and projected to the Attributor dimension:
\begin{equation}
    \widetilde Z_c
    =
    \operatorname{Proj}_E
    \left(
    \operatorname{Flatten}(Z_c)
    \right).
\end{equation}

The entity queries retrieve evidence from the current representation through cross-attention:
\begin{equation}
    A_E
    =
    \operatorname{CrossAttn}
    \left(
    \operatorname{LN}(Q_E),
    \operatorname{LN}(\widetilde Z_c)
    \right).
    \label{eq:app-entity-attn}
\end{equation}
After the attribution blocks, the resulting entity states are
\begin{equation}
    S^E = F_E(Z_c).
    \label{eq:app-entity-output}
\end{equation}

Thus, the Entity Attributor operates only on the current representation. The predicted future does not enter the hierarchy at this stage.

\subsubsection{Dynamic Attributor}

The Dynamic Attributor converts each Entity state into a temporally resolved representation, allowing predicted changes to be attributed to individual entities. For entity $i$ at temporal position $t$, we initialize an entity-conditioned temporal query as
\begin{equation}
q_{i,t}^{(0)}
=
\phi_E(S_i^E)
+
e_t
+
e_{\mathrm{src}(t)},
\label{eq:app-dynamic-query}
\end{equation}
where $\phi_E(S_i^E)$ carries the identity of entity $i$, $e_t$ encodes the temporal position, and $e_{\mathrm{src}(t)}$ indicates whether the position corresponds to the observed context or the predicted future. Collecting all queries gives $Q_D^{(0)}\in\mathbb{R}^{K\times T\times d_D}$.

The entity-conditioned queries then retrieve evidence jointly from the observed and predicted-future representations:
\begin{equation}
H_D
=
\operatorname{CrossAttn}
\left(
Q_D^{(0)},
\operatorname{Proj}_D
\left([Z_c;\widehat Z_f]\right)
\right).
\label{eq:app-dynamic-attn}
\end{equation}
Since each query is tied to a specific Entity state, the retrieved current--future evidence is attributed to that entity rather than pooled into a global temporal representation.

Temporal dependencies are then modeled independently along each entity trajectory:
\begin{equation}
\widetilde S^D_{i,1:T}
=
\operatorname{TemporalTransformer}
\left(
H_{D,i,1:T}
\right).
\label{eq:app-dynamic-temporal}
\end{equation}
This forms the main attribution path, where $\widetilde S^D$ captures entity-specific temporal changes inferred from both the observed state and its predicted future.

In parallel, we introduce a residual information path
\begin{equation}
S^D_{\mathrm{res}}
=
G_D(S^E,Z_c),
\label{eq:app-dynamic-residual}
\end{equation}
where $G_D$ maps the Entity states together with the current visual representation into the same entity--time feature space as $\widetilde S^D$. Unlike the main attribution path, $G_D$ does not access $\widehat Z_f$. Its role is to preserve entity identity and current-state evidence that may not be expressed as temporal change.

The final Dynamic state combines the two paths:
\begin{equation}
S^D
=
\widetilde S^D
+
S^D_{\mathrm{res}}.
\label{eq:app-dynamic}
\end{equation}
Thus, $S^D\in\mathbb{R}^{K\times T\times d_D}$ retains explicit entity and temporal axes: each $S^D_{i,t}$ contains both the future-conditioned change attributed to entity $i$ and the lower-level current-state information preserved by the residual path.

\subsubsection{Relation Attributor}

The Relation Attributor converts entity-specific dynamics into explicit pairwise interaction states. We consider all unordered entity pairs
$\mathcal{P}={(i,j)\mid 1\leq i<j\leq K}$.
For each pair $(i,j)$ at temporal position $t$, we construct a pair representation from their Entity and Dynamic states:
\begin{equation}
r_{ij,t}
=
\left[
S_i^E + S_j^E,\;
|S_i^E-S_j^E|,\;
S_{i,t}^D + S_{j,t}^D,\;
|S_{i,t}^D-S_{j,t}^D|
\right]
\label{eq:app-relation-input}
\end{equation}
The sum terms capture information shared by the two entities, while the difference terms encode their relative entity and dynamic states. Using symmetric operations also makes $r_{ij,t}$ invariant to the ordering of the pair.

Each pair representation is first projected into the Relation feature space and then aggregated along its temporal trajectory:
\begin{equation}
\widetilde S^R_{ij,1:T_R}
=
\operatorname{TemporalAttention}
\left(
\operatorname{MLP}_R
\left(
r_{ij,1:T_R}
\right)
\right).
\label{eq:app-relation-temporal}
\end{equation}
This forms the main relation-attribution path, where $\widetilde S^R_{ij,t}$ represents the interaction evidence associated with entity pair $(i,j)$ at temporal position $t$.

In parallel, we introduce a residual information path
\begin{equation}
S^R_{\mathrm{res}}
=
G_R(S^D,Z_c),
\label{eq:app-relation-residual}
\end{equation}
where $G_R$ maps the Dynamic states together with the current visual representation into the same pair--time feature space as $\widetilde S^R$. This path preserves lower-level dynamic and current-state evidence that may not be fully retained by the explicit pairwise transformation.

The final Relation state is obtained by combining the two paths:
\begin{equation}
S^R
=
\widetilde S^R
+
S^R_{\mathrm{res}}.
\label{eq:app-relation}
\end{equation}
Importantly, the Relation Attributor does not independently attend to $\widehat Z_f$. Future-conditioned evidence has already been attributed to individual entities in $S^D$ and is therefore propagated naturally into pairwise reasoning. The resulting $S^R$ preserves explicit pair and temporal axes, with each $S^R_{ij,t}$ describing the attributed interaction state of entity pair $(i,j)$ at temporal position $t$.

\subsection{Hierarchical State Construction}
\label{app:hierarchical-state}

The complete HASP computation follows the sequential attribution structure:
\begin{align}
    S^E
    &= F_E(Z_c), \\
    S^D
    &= F_D(S^E,Z_c,\widehat Z_f)
    + G_D(S^E,Z_c), \\
    S^R
    &= F_R(S^E,S^D)
    + G_R(S^D,Z_c).
    \label{eq:app-hasp}
\end{align}

Accordingly, the native representation axis changes progressively as
\begin{equation}
    (T_c,N)
    \rightarrow
    (K)
    \rightarrow
    (K,T_D)
    \rightarrow
    (|\mathcal P|,T_R).
\end{equation}

This sequential organization distinguishes HASP from independent feature heads: each Attributor consumes the structured state produced by the preceding level.

\subsection{Structured State and Downstream Readout}
The three levels jointly form the abductive world state
\begin{equation}
    S=(S^E,S^D,S^R).
\end{equation}
Importantly, the levels are complementary rather than interchangeable. $S^E$ provides object-centered identity and spatial evidence, $S^D$ binds predicted temporal change to individual entities, and $S^R$ represents temporally evolving pairwise interactions. Higher levels introduce additional structural organization while residual paths preserve information established at lower levels.

For a downstream task $\tau$ with target $Y_\tau$, a lightweight task-specific readout operates on the state:
\begin{equation}
    \widehat{Y}_{\tau}
    =
    R_{\tau}(S^E,S^D,S^R).
    \label{eq:task-readout}
\end{equation}
This separation allows us to evaluate whether a structured abductive state provides a more useful interface for reasoning than the original predictive latent representation without modifying the predictive backbone itself.

\subsection{Factor-specific Training Objectives}
\label{app:training-detail}

HASP is optimized with factor-specific objectives defined at the native granularity of each state:
\begin{equation}
    \mathcal L_{\mathrm{HASP}}
    =
    \mathcal L_E
    +
    \mathcal L_D
    +
    \mathcal L_R
    +
    \lambda_{\mathrm{task}}\mathcal L_{\mathrm{task}}.
    \label{eq:app-training-loss}
\end{equation}

Here, $\mathcal L_E$ supervises Entity states, $\mathcal L_D$ supervises entity-time Dynamic states, and $\mathcal L_R$ supervises entity-pair-time Relation states. When a downstream task is jointly optimized, $\mathcal L_{\mathrm{task}}$ provides the corresponding task-level supervision.

For the structured probe used in Physion++, the factor-specific objectives are instantiated over the corresponding object, temporal, and pairwise annotations. Invalid object, time, or pair entries are excluded from the respective losses.

Throughout HASP training, the pretrained encoder $E$ and latent predictor $P$ remain frozen. The predictor is trained separately to provide the current-to-future latent prediction and is subsequently fixed during HASP optimization.

\subsection{Complete Inference Procedure}
\label{app:inference-procedure}

The complete AWM inference procedure can be summarized as follows.

\begin{quote}
\textbf{Algorithm AWM (Abductive World Modeling).}
Given observed video context $X_{1:t}$, first compute
$Z_c\gets E(X_{1:t})$ and
$\widehat Z_f\gets P(Z_c)$.
Then construct Entity states
$S^E\gets F_E(Z_c)$
from the current representation. Next, use the Entity states together with the current and predicted-future representations to construct the Dynamic states
$S^D\gets F_D(S^E,Z_c,\widehat Z_f)+G_D(S^E,Z_c)$.
Finally, enumerate the unordered entity pairs
$\mathcal P=\{(i,j):i<j\}$,
construct pair representations from the Entity and Dynamic states, and obtain
$S^R\gets F_R(S^E,S^D)+G_R(S^D,Z_c)$.
The resulting hierarchical state
$S=(S^E,S^D,S^R)$
is passed to the downstream readout
$\widehat Y_\tau\gets R_\tau(S)$.
\end{quote}

\section{Additional Results}
\label{app:experiment_detail}

\subsection{Object Identity Diagnostics}
\label{app:attributor-analysis}

We further examine whether the Entity state forms an object-centered representation rather than a pooled scene feature. Beyond the native-factor probes reported in the main text, we evaluate slot occupancy, identity consistency, object attributes, and geometric properties on Physion++.

\input{Table/physion_slot_diagnostics}

As shown in Table~\ref{tab:physion-slot-diagnostics}, Entity slots exhibit strong object-level semantics, achieving 99.77\% presence F1, 95.79\% target-object accuracy, and 92.86\% object-type accuracy. Identity consistency reaches 70.29\%, with an ID switch rate of 4.05\%, indicating that object identity is substantially preserved across time. The low center, geometry, and velocity errors further show that individual slots retain spatial and motion information associated with their corresponding objects. These diagnostics complement the native-factor results in the main text by providing a more detailed characterization of the object-centered structure of the Entity state.

\subsection{Additional Entity Intervention Diagnostics}
\label{app:entity-intervention-diagnostics}

We further analyze the Physion++ Entity intervention from Experiment~3 by examining how target-object masking changes the OCP decision pattern. In addition to the aggregate accuracy and AUROC results reported in the main text, we report confusion counts and paired prediction-logit changes.

\input{Table/physion_confusion_intervention}

Table~\ref{tab:physion-confusion-intervention} shows that target masking produces a systematic change in the prediction boundary. Relative to the full Visual + ShallowProbe condition, the number of true positives increases from 277 to 325, while true negatives decrease from 297 to 181 and false positives increase from 124 to 240. Thus, removing the target Entity slot does not simply suppress positive evidence; instead, it substantially alters how the readout separates positive and negative outcomes. In comparison, irrelevant-slot masking and random-slot intervention produce much smaller changes in the confusion pattern.

\input{Table/physion_logit_intervention}

The paired-logit analysis in Table~\ref{tab:physion-logit-intervention} further quantifies this sensitivity. Target masking produces a mean absolute logit change of 0.67680, approximately 3.9 times that of irrelevant masking (0.17355) and 7.6 times that of random-slot intervention (0.08895). These results provide additional evidence that OCP prediction is selectively sensitive to the Entity representation associated with the target object rather than to arbitrary perturbations of the slot representation.

\subsection{CLEVRER Relevant-Object Intervention}
\label{app:query-object-intervention}

We additionally evaluate whether predictive reasoning on CLEVRER depends selectively on objects that are relevant to the current query. The evaluation contains 3,557 predictive questions and 7,114 answer options. We compare masking the query-relevant object with masking an irrelevant object while keeping the model and task readout fixed.

\input{Table/query_intervention}

As shown in Table~\ref{tab:query-intervention}, masking the relevant object reduces question accuracy from 50.18 to 43.46 and option accuracy from 71.68 to 66.80. In comparison, masking an irrelevant object retains higher performance, with 47.34 question accuracy and 69.55 option accuracy.

The larger degradation under relevant-object masking shows that the performance drop is not caused merely by removing an arbitrary object. Instead, the CLEVRER readout is more sensitive to Entity evidence associated with the current query, providing additional evidence that object-level information is used selectively during downstream reasoning.

\section{Details of Assets Used in This Paper}
\label{app:dataset_detail}

In our experiments, we evaluate AWM on three video benchmarks spanning physical prediction, causal reasoning, and action understanding. We follow the corresponding benchmark settings and use the same data splits for all compared methods.

\subsection{Physical Prediction Dataset}

\textbf{Physion++}~\citep{bear2021physion} is a benchmark for evaluating physical prediction from videos of interacting objects. The scenes contain diverse physical configurations and interactions, requiring models to reason about object dynamics and predict future physical outcomes. In our experiments, Physion++ is used to evaluate whether the learned representation captures physical information relevant to future evolution.

\subsection{Causal Reasoning Dataset}

\textbf{CLEVRER}~\citep{yi2020clevrer} is a synthetic video reasoning benchmark designed to evaluate understanding of objects, motion, collisions, and temporal events. It contains questions that require reasoning about observed and future events based on the dynamics and interactions among objects. In our experiments, CLEVRER is used to evaluate whether AWM captures Entity, Dynamic, and Relation structure that supports causal reasoning about video events.

\subsection{Action Understanding Dataset}

\textbf{EPIC-KITCHENS-100 (EK100)}~\citep{damend2022epic} is a large-scale egocentric video benchmark containing unconstrained first-person recordings of everyday activities. Each action segment is annotated with a verb and noun pair, together defining an action class. In our experiments, EK100 is used to evaluate verb, noun, and action recognition, providing a substantially different setting from the synthetic and physics-oriented benchmarks above.

\subsection{Data Splits and Evaluation Settings}

For each benchmark, we follow its corresponding experimental protocol and use the same training and evaluation splits across all compared methods. Physion++ is evaluated for physical prediction, CLEVRER for causal reasoning, and EK100 for verb, noun, and action recognition. For CLEVRER, we evaluate only on the predictive subset of the test set, which contains questions requiring prediction of future events.

\section{Details of Methods and Experimental Settings}
\label{appendix:exp_details}

\begin{table*}[ht]
    \centering
    \caption{
    \textbf{Main configuration and hyperparameter settings of AWM.}
    We report the predictive backbone, evaluation configuration, and computational setup used in our experiments.
    }
    \label{tab:parameter_setting}

    \footnotesize
    \renewcommand{\arraystretch}{1.08}

    \begin{tabular*}{0.72\linewidth}{
        @{\extracolsep{\fill}}
        l
        c
        @{}
    }
        \toprule\toprule
        \textbf{Parameter} & \textbf{Value} \\
        \midrule

        \multicolumn{2}{c}{\textbf{Predictive Backbone}} \\
        \midrule
        Backbone                      & V-JEPA 2 ViT-H \\
        Observed frames               & 16 \\
        Predicted future frames       & 16 \\
        Patch size                    & 16 \\
        Tubelet size                  & 2 \\
        Predictor depth               & 12 \\
        Predictor embedding dimension & 384 \\
        Predictor attention heads     & 12 \\

        \midrule
        \multicolumn{2}{c}{\textbf{Evaluation Configuration}} \\
        \midrule
        Visual encoder        & Frozen \\
        Latent predictor      & Frozen \\
        HASP state extractor  & Frozen \\
        Trainable modules     & Factor probes / task readouts \\
        Readout optimizer     & AdamW \\
        Precision             & BF16 \\
        Random seed           & 239 \\

        \bottomrule
    \end{tabular*}

\end{table*}

\subsection{Details of Experimental Setup}
\label{appendix:exp_setup}

In this section, we provide additional details about the implementation of AWM and the baselines, together with the evaluation settings used in our experiments.

\textbf{Implementation Details of the Baselines.}
We compare AWM with V-JEPA 2, Orca, and VideoMAE~v2. Within each benchmark, we use the same data splits and evaluation metrics whenever the corresponding model supports the required protocol. AWM and V-JEPA 2 share the same predictive backbone and input setting, providing a matched comparison for evaluating the contribution of abductive world modeling.

\textbf{V-JEPA 2}
is the matched predictive baseline of AWM. It uses the same ViT-H encoder and latent predictor as AWM to encode the observed context and predict future latent states. Downstream readouts operate directly on the V-JEPA 2 representations without introducing explicit Entity, Dynamic, or Relation states. This comparison therefore isolates the effect of HASP under the same predictive backbone.

\textbf{Orca}
is an external video representation baseline. Since Orca does not provide a future predictor compatible with our evaluation pipeline, it cannot directly generate the predicted future latent representation required by AWM. On Physion++, where ground-truth future frames are available, we therefore encode both the current and future videos with the frozen Orca encoder and use the resulting future representation as a substitute for the predicted future latent. On EK100, the Orca encoder is kept frozen and only a lightweight attentive readout is trained for verb, noun, and action recognition.

\textbf{VideoMAE~v2}
is another external video representation baseline. As VideoMAE~v2 also does not provide a compatible future predictor, we use the same substitution on Physion++ by encoding the available ground-truth future frames to obtain the future latent representation. On EK100, unlike Orca, the pretrained VideoMAE~v2 encoder is fine-tuned together with the task-specific verb, noun, and action classification heads.

Because the predictive subset of CLEVRER requires reasoning about unobserved future events and does not provide the corresponding ground-truth future frames, this substitution is not available for Orca or VideoMAE~v2. We therefore do not report CLEVRER results for these two methods rather than approximating the missing predicted future representation.

\textbf{AWM}
is our proposed framework and is built on the same V-JEPA 2 predictive backbone used by the matched baseline. Given an observed video context, the frozen encoder produces the current representation $Z_c$, and the frozen latent predictor generates the predicted future representation $\widehat{Z}_f$ without accessing ground-truth future frames. HASP then performs backward abductive inference over these representations to construct the Entity, Dynamic, and Relation states. During downstream evaluation, the V-JEPA 2 encoder, predictor, and HASP are kept frozen, and only lightweight factor probes or task-specific readouts are trained.

\textbf{Hyperparameters.}
Table~\ref{tab:parameter_setting} summarizes the main configuration of AWM used throughout our experiments. Unless otherwise specified, we use the same frozen representation for downstream evaluation and train independent readouts for physical prediction, causal reasoning, and action understanding.

\subsection{Algorithm of Abductive World Modeling}
\label{appendix:algorithm}

We summarize the inference procedure of AWM in Algorithm~\ref{alg:awm}. AWM first predicts a future latent state from the observed video context and then performs backward abductive inference through HASP. The Entity Attributor identifies what exists, the Dynamic Attributor infers how the identified entities change using the predicted future, and the Relation Attributor infers how pairs of entities interact. These factors jointly form the structured causal state used for downstream prediction and reasoning.

\begin{algorithm}[t]
\caption{Abductive World Modeling}
\label{alg:awm}
\begin{algorithmic}[1]

\Require Observed video context $X_{1:t}$
\Require Frozen visual encoder $E$ and latent predictor $P$
\Require Entity, Dynamic, and Relation Attributors $F_E$, $F_D$, and $F_R$
\Require Task-specific readout $R_{\tau}$

\State Encode the observed context: $Z_c \gets E(X_{1:t})$
\State Predict the future latent state: $\widehat{Z}_f \gets P(Z_c)$
\State \textit{Abduce backward through HASP:}
\State Infer the Entity state: $S^E \gets F_E(Z_c)$
\State Infer the Dynamic state: $S^D \gets F_D(S^E,Z_c,\widehat{Z}_f)+G_D(S^E,Z_c)$
\State Construct candidate entity pairs: $\mathcal{P}\gets\{(i,j)\mid i<j\}$
\For{each $(i,j)\in\mathcal{P}$}
    \For{$t=1,\ldots,T_R$}
        \State Construct pairwise Entity and Dynamic features:
        \Statex $r_{ij,t}\gets \left[S_i^E+S_j^E,\; |S_i^E-S_j^E|,\; S_{i,t}^D+S_{j,t}^D,\; |S_{i,t}^D-S_{j,t}^D|\right]$
    \EndFor
\EndFor
\State Infer the Relation state: $S^R \gets F_R(\{r_{ij,1:T_R}\}_{(i,j)\in\mathcal{P}})+G_R(S^D,Z_c)$
\State Construct the unified abductive state: $S\gets(S^E,S^D,S^R)$
\State Produce the downstream prediction: $\widehat{Y}_{\tau}\gets R_{\tau}(S)$

\State \Return $S,\widehat{Y}_{\tau}$

\end{algorithmic}
\end{algorithm}

\end{document}

%% file: math_commands.tex
%%%%% NEW MATH DEFINITIONS %%%%%

\usepackage{amsmath,amsfonts,bm}

% Mark sections of captions for referring to divisions of figures

% Highlight a newly defined term

% Figure reference, lower-case.

% Figure reference, capital. For start of sentence

% Section reference, lower-case.

% Section reference, capital.

% Reference to two sections.

% Reference to three sections.

% Reference to an equation, lower-case.
\def\eqref#1{equation~\ref{#1}}
% Reference to an equation, upper case

% A raw reference to an equation---avoid using if possible

% Reference to a chapter, lower-case.

% Reference to an equation, upper case.

% Reference to a range of chapters

% Reference to an algorithm, lower-case.

% Reference to an algorithm, upper case.

% Reference to a part, lower case

% Reference to a part, upper case

\def\1{\bm{1}}

% Random variables

% rm is already a command, just don't name any random variables m

% Random vectors

% Elements of random vectors

% Random matrices

% Elements of random matrices

% Vectors

% Elements of vectors

% Matrix

% Tensor
\DeclareMathAlphabet{\mathsfit}{\encodingdefault}{\sfdefault}{m}{sl}
\SetMathAlphabet{\mathsfit}{bold}{\encodingdefault}{\sfdefault}{bx}{n}

% Graph

% Sets

% Don't use a set called E, because this would be the same as our symbol
% for expectation.

% Entries of a matrix

% entries of a tensor
% Same font as tensor, without \bm wrapper

% The true underlying data generating distribution

% The empirical distribution defined by the training set

% The model distribution

% Stochastic autoencoder distributions

 % Laplace distribution

% Wolfram Mathworld says $L^2$ is for function spaces and $\ell^2$ is for vectors
% But then they seem to use $L^2$ for vectors throughout the site, and so does
% wikipedia.

 % See usage in notation.tex. Chosen to match Daphne's book.

%% file: Table/main_results.tex
\begin{table}[t]
\centering
\caption{Main comparison across datasets. All entries are percentage values reported without the percent sign, and the best value in each metric column is boldfaced. ``Bal.~Acc.'' and ``Acc.'' denote balanced accuracy and accuracy, respectively, while EK100 results are reported as Top-1/Top-5. Orca and VideoMAE~v2 are marked as ``\textemdash'' on CLEVRER because their released models do not provide the predictive module required by the CLEVRER evaluation pipeline.}
\label{tab:main}
\scriptsize
\setlength{\tabcolsep}{3.2pt}
\renewcommand{\arraystretch}{0.92}
\resizebox{\textwidth}{!}{
\begin{tabular}{@{}lccc cc ccc@{}}
\toprule\toprule
 & \multicolumn{3}{c}{\textbf{Physion++}} & \multicolumn{2}{c}{\textbf{CLEVRER}} & \multicolumn{3}{c}{\textbf{EK100 (T1/T5)}} \\
\cmidrule(lr){2-4}\cmidrule(lr){5-6}\cmidrule(lr){7-9}
\textbf{Method} & \textbf{AUROC} & \textbf{Bal.~Acc.} & \textbf{Acc.} & \textbf{Option Acc.} & \textbf{Question Acc.} & \textbf{Verb} & \textbf{Noun} & \textbf{Action} \\
\midrule
\vjepa{} & $65.20$ & $61.57$ & $61.60$ & $65.46$ & $42.20$ & $55.74/86.81$ & $34.65/62.67$ & $25.67/46.94$ \\
Orca & $59.15$ & $55.94$ & $56.64$ & \textemdash & \textemdash & $33.75/74.39$ & $21.54/45.36$ & $13.08/30.46$ \\
VideoMAE~v2 & $67.61$ & $62.60$ & $62.86$ & \textemdash & \textemdash & $59.54/82.57$ & $46.37/68.79$ & $35.96/54.57$ \\
\awm{} (Ours) & \best{72.19} & \best{66.17} & \best{65.63} & \best{70.89} & \best{49.31} & \best{69.05/91.27} & \best{47.59/75.05} & \best{43.12/67.34} \\
\bottomrule
\end{tabular}
}
\end{table}

%% file: Table/input_temporal_intervention.tex
\begin{table}[t]
\centering
\caption{
Temporal-static latent intervention on Physion++.
\textit{AWM} uses the original context and future features.
\textit{AWM w/o Temporal} temporally averages and repeats these features at each time step, removing temporal variation while keeping the model and readouts frozen.
}
\label{tab:input-temporal-intervention}
\small
\renewcommand{\arraystretch}{1.03}

\begin{tabularx}{\linewidth}{
@{}
>{\raggedright\arraybackslash}X
>{\centering\arraybackslash}X
>{\centering\arraybackslash}X
>{\centering\arraybackslash}X
@{}
}
\toprule\toprule
\textbf{Condition}
& \textbf{Speed MAE} ($\downarrow$)
& \textbf{Contact AUROC} ($\uparrow$)
& \textbf{OCP AUROC} ($\uparrow$) \\
\midrule

\textbf{AWM}
& \textbf{0.0670}
& \textbf{0.9837}
& \textbf{0.7907} \\

w/o Temporal
& 0.1120
& 0.9485
& 0.5646 \\

\bottomrule
\end{tabularx}
\end{table}

%% file: Table/relation_selective_intervention.tex
\begin{table}[t]
\centering
\caption{
Relation-pair intervention on CLEVRER.
\textit{AWM} uses all Relation pairs;
\textit{AWM w/o Contact Pair} removes the target contact pair;
and \textit{AWM w/o Non-contact Pair} removes an unrelated pair.
We report geometric and interaction prediction performance.
}
\label{tab:relation-selective-intervention}
\small
\renewcommand{\arraystretch}{1.05}

\begin{tabularx}{\linewidth}{
@{}
>{\raggedright\arraybackslash}X
>{\centering\arraybackslash}X
>{\centering\arraybackslash}X
>{\centering\arraybackslash}X
@{}
}
\toprule\toprule
\textbf{Condition}
& \textbf{Pair-dist. MAE} ($\downarrow$)
& \textbf{TTC MAE} ($\downarrow$)
& \textbf{Contact AUROC} ($\uparrow$) \\
\midrule

\textbf{AWM}
& \textbf{0.0196}
& \textbf{0.1306}
& \textbf{0.9633} \\

w/o Contact Pair
& 0.0225
& 0.1376
& 0.9488 \\

w/o Non-contact Pair
& 0.0208
& 0.1349
& 0.9518 \\

\bottomrule
\end{tabularx}
\end{table}

%% file: Table/physion_slot_diagnostics.tex
\begin{table}[t]
\centering
\caption{Additional diagnostics of the Physion++ Entity slots.}
\label{tab:physion-slot-diagnostics}

\footnotesize
\renewcommand{\arraystretch}{1.08}

\begin{tabular*}{0.72\linewidth}{@{\extracolsep{\fill}}lc@{}}
\toprule\toprule
\textbf{Diagnostic} & \textbf{Value} \\
\midrule
Presence F1             & 99.77   \\
Slot occupancy          & 64.52   \\
Identity consistency    & 70.29   \\
ID switch rate          & 4.05    \\
Target-object accuracy  & 95.79   \\
Object-type accuracy    & 92.86   \\
Color RGB MAE           & 0.2244  \\
Center MAE              & 0.0454  \\
Geometry MAE            & 0.0285  \\
Velocity MAE            & 0.00629 \\
\bottomrule
\end{tabular*}

\end{table}

%% file: Table/physion_confusion_intervention.tex
\begin{table}[t]
\centering
\caption{Physion++ OCP confusion counts under probe interventions.}
\label{tab:physion-confusion-intervention}

\footnotesize
\renewcommand{\arraystretch}{1.08}

\begin{tabular*}{0.72\linewidth}{@{\extracolsep{\fill}}lrrrr@{}}
\toprule\toprule
\textbf{Condition} & \textbf{TP} & \textbf{TN} & \textbf{FP} & \textbf{FN} \\
\midrule
Visual-only           & 237 & 292 & 129 & 142 \\
Visual + ShallowProbe & 277 & 297 & 124 & 102 \\
Target mask           & 325 & 181 & 240 & 54  \\
Irrelevant mask       & 262 & 305 & 116 & 117 \\
Random slot           & 279 & 294 & 127 & 100 \\
\bottomrule
\end{tabular*}

\end{table}

%% file: Table/physion_logit_intervention.tex
\begin{table}[t]
\centering
\caption{Paired OCP-logit changes under probe-level interventions.}
\label{tab:physion-logit-intervention}

\footnotesize
\renewcommand{\arraystretch}{1.08}

\begin{tabular*}{0.72\linewidth}{@{\extracolsep{\fill}}lcc@{}}
\toprule\toprule
\textbf{Condition}
& \textbf{Mean change}
& \textbf{Mean absolute change} \\
\midrule
Target mask     & +0.49195 & 0.67680 \\
Irrelevant mask & -0.08051 & 0.17355 \\
Random slot     & +0.02437 & 0.08895 \\
\bottomrule
\end{tabular*}

\end{table}

%% file: Table/query_intervention.tex
\begin{table}[t]
\centering
\caption{
Relevant-object intervention on CLEVRER predictive queries.
\textit{AWM} uses all Entity evidence;
\textit{AWM w/o Relevant Object} removes the object referred to by the query;
and \textit{AWM w/o Irrelevant Object} removes an object unrelated to the query.
}
\label{tab:query-intervention}
\small
\renewcommand{\arraystretch}{1.03}

\begin{tabularx}{0.78\linewidth}{@{}Lcccc@{}}
\toprule\toprule
\textbf{Condition}
& \makecell{\textbf{Question}\\\textbf{Correct}}
& \makecell{\textbf{Question}\\\textbf{Acc.} $\uparrow$}
& \makecell{\textbf{Option}\\\textbf{Correct}}
& \makecell{\textbf{Option}\\\textbf{Acc.} $\uparrow$} \\
\midrule
\textbf{AWM}
& 1,785/3,557
& \textbf{50.18\%}
& 5,099/7,114
& \textbf{71.68\%} \\

AWM w/o Relevant Object
& 1,546/3,557
& 43.46\%
& 4,752/7,114
& 66.80\% \\

AWM w/o Irrelevant Object
& 1,684/3,557
& 47.34\%
& 4,948/7,114
& 69.55\% \\
\bottomrule
\end{tabularx}
\end{table}